\documentclass[conference]{IEEEtran}
\IEEEoverridecommandlockouts

\usepackage{booktabs, multirow}
\usepackage[hidelinks]{hyperref}
\usepackage{cite}
\usepackage{amsmath, amssymb, amsfonts}
\usepackage{algorithmic}
\usepackage{graphicx}
\usepackage{textcomp}
\usepackage{xcolor}
\def\BibTeX{{\rm B\kern-.05em{\sc i\kern-.025em b}\kern-.08em
    T\kern-.1667em\lower.7ex\hbox{E}\kern-.125emX}}

\makeatletter
\newcommand{\linebreakand}{
	\end{@IEEEauthorhalign}
	\hfill\mbox{}\par
	\mbox{}
	\hfill
	\begin{@IEEEauthorhalign}
}
\makeatother

\begin{document}

\title{MSVM-UNet: Multi-Scale Vision Mamba UNet for Medical Image Segmentation}

\author{
	\IEEEauthorblockN{Chaowei Chen}
	\IEEEauthorblockA{\textit{School of Information Science and Engineering} \\
	\textit{Yunnan University} \\ Kunming, China \\ \href{mailto:chishengchen@stu.ynu.edu.cn}{chishengchen@stu.ynu.edu.cn}}
	\and
	\IEEEauthorblockN{Li Yu}
	\IEEEauthorblockA{\textit{School of Information Science and Engineering} \\
	\textit{Yunnan University} \\ Kunming, China \\ \href{mailto:yuli0501@163.com}{yuli0501@163.com}}
	\linebreakand
	\IEEEauthorblockN{Shiquan Min}
	\IEEEauthorblockA{\textit{School of Information Science and Engineering} \\
	\textit{Yunnan University} \\ Kunming, China \\ \href{mailto:minshiquan@mail.ynu.edu.cn}{minshiquan@mail.ynu.edu.cn}}
	\and
	\IEEEauthorblockN{Shunfang Wang* \thanks{* corresponding author.}}
	\IEEEauthorblockA{\textit{School of Information Science and Engineering} \\
	\textit{Yunnan University} \\ Kunming, China \\ \href{mailto:sfwang_66@ynu.edu.cn}{sfwang\_66@ynu.edu.cn}}
}

\maketitle

\begin{abstract}
State Space Models (SSMs), especially Mamba, have shown great promise in medical image segmentation due to their ability to model long-range dependencies with linear computational complexity. However, accurate medical image segmentation requires the effective learning of both multi-scale detailed feature representations and global contextual dependencies. Although existing works have attempted to address this issue by integrating CNNs and SSMs to leverage their respective strengths, they have not designed specialized modules to effectively capture multi-scale feature representations, nor have they adequately addressed the directional sensitivity problem when applying Mamba to 2D image data. To overcome these limitations, we propose a Multi-Scale Vision Mamba UNet model for medical image segmentation, termed MSVM-UNet. Specifically, by introducing multi-scale convolutions in the VSS blocks, we can more effectively capture and aggregate multi-scale feature representations from the hierarchical features of the VMamba encoder and better handle 2D visual data. Additionally, the large kernel patch expanding (LKPE) layers achieve more efficient upsampling of feature maps by simultaneously integrating spatial and channel information. Extensive experiments on the Synapse and ACDC datasets demonstrate that our approach is more effective than some state-of-the-art methods in capturing and aggregating multi-scale feature representations and modeling long-range dependencies between pixels. Our implementation is available at \href{https://github.com/gndlwch2w/msvm-unet}{\textit{https://github.com/gndlwch2w/msvm-unet}}.
\end{abstract}


\section{Introduction}
Precise and efficient medical image segmentation is one of the fundamental yet challenging tasks in medical image analysis. Research in this area leverages techniques such as deep learning to analyze various types of medical images and produce segmentation maps for specific organs or pathological regions, thereby aiding physicians and researchers in disease analysis and diagnosis.

In recent years, medical image segmentation with Convolutional Neural Networks (CNNs) and Vision Transformers (ViTs) has seen notable success. Specifically, UNet \cite{ronneberger2015u}, with its elegant U-shaped structure and skip connections, excels in processing high-resolution medical images and seamlessly combines low-level details with high-level semantics for impressive segmentation results. Furthermore, TransUNet \cite{chen2021transunet} proposes a hybrid structure of CNNs and ViTs to simultaneously utilize the detail extraction capabilities of CNNs and the long-range dependency modeling capabilities of ViTs. Although these methods achieve commendable performance and produce high-quality segmentation results, inherent characteristics of CNNs and transformers present performance bottlenecks \cite{chen2021transunet, rahman2023medical}. Specifically, CNNs rely on local convolutional kernels for feature extraction, which, while effective for capturing local feature patterns, limits their ability to model global and geometric features \cite{zhang2019self}. Although transformer-based methods perform well in modeling long-range dependencies, self-attention mechanisms have quadratic computational complexity with respect to sequence length \cite{dosovitskiy2020image}, which makes it challenging to handle high-resolution segmentation tasks efficiently. Moreover, methods such as Swin Transformer \cite{liu2021swin}, PVT v2 \cite{wang2022pvt}, and BiFormer \cite{zhu2023biformer} propose effective self-attention computation techniques. However, these methods come with a trade-off between computational complexity and modeling capability, thereby limiting their ability to model long sequences.

Recently, State Space Models (SSMs) \cite{gu2021efficiently, mehta2022long, gu2023mamba} have garnered widespread research interest due to their immense potential in modeling long sequences. Mamba \cite{gu2023mamba} is proficient in modeling long-sequence dependencies with linear computational complexity and has demonstrated significant success in the field of natural language processing. Building on this, VMamba \cite{liu2024vmamba} introduces a Cross-Scan Module (CSM) and a well-designed hierarchical architecture design, indicating its significant potential in analyzing 2D image data. In the field of medical image segmentation, efficiently processing high-resolution medical images remains a significant challenge. Inspired by the above work, U-Mamba \cite{ma2024u} proposes embedding convolutional operations within SSMs to integrate the local feature extraction power of convolutional layers with the long-range dependency capture capabilities of SSMs. Swin-UMamba \cite{liu2024swin} demonstrates that transferring VMamba pre-trained on ImageNet-1k to the medical image segmentation domain can effectively address issues such as limited data resources. Similar to Swin-UNet \cite{cao2022swin}, VM-UNet \cite{ruan2024vm} proposes using pure Visual State Space (VSS) blocks to construct a medical image segmentation framework.

Unlike 1D sequences, pixels in 2D visual data inherently possess directional dependencies \cite{yang2023directional}. Directly applying the 1D sequence processing method from Mamba to 2D data fails to effectively capture long-range dependencies between pixels and results in limited receptive fields, due to the lack of consideration for the spatial characteristics of 2D data, which is known as the directional sensitivity problem \cite{liu2024vmamba}. Although VMamba employs four scanning strategies to address this issue, it only focuses on four of the eight neighboring directions (i.e., up, down, left, right), leading to certain limitations in modeling dependencies between pixels due to the incomplete consideration of dependencies in all directions. Additionally, although U-Mamba and Swin-UMamba incorporate a hybrid architecture of CNNs and SSMs, they do not specifically address multi-scale feature learning, which results in shortcomings when analyzing objects of varying sizes and shapes \cite{heidari2023hiformer, chen2021crossvit}. To solve these issues, we propose a Multi-Scale Visual State Space (MSVSS) block, which uses a set of parallel convolution operations with different kernel sizes to capture and aggregate multi-scale feature representations, and not only models dependencies in the original four directions, but also uses convolution operations to aggregate the information in the remaining four diagonal directions.

Moreover, the patch expanding layer is used for feature upsampling in both Swin-UNet and VM-Net. However, since the patch expanding layer only considers channel-wise information and does not account for spatial relationships during upsampling, it leads to insufficient discriminative power. To overcome this issue, we propose a Large Kernel Patch Expanding (LKPE) layer for upsampling, which aggregates features along the channel dimension while also integrating spatial information through a $3 \times 3$ depth-wise convolution.

The main contributions of this study are summarized as follows:
\begin{itemize}
	\item We propose a novel Multi-Scale Visual State Space (MSVSS) block, which combines CSM with multi-scale convolution operations to not only effectively model long-range dependencies between pixels but also capture multi-scale feature representations.
	
	\item We introduce a new Large Kernel Patch Expanding (LKPE) layer for feature map upsampling. By incorporating a large kernel depth-wise convolution before expanding the channel dimensions, we achieve more discriminative feature representations with acceptable additional overhead.
	
	\item We validate our proposed MSVM-Net on the Synapse multi-organ dataset and the ACDC dataset. Specifically, on the Synapse multi-organ dataset, our model achieved a DSC of 85.00\% and an HD95 of 14.75mm. On the ACDC dataset, our model achieved a DSC of 92.58\%.
\end{itemize} 

\section{METHODS}
\subsection{Overall Architecture of MSVM-UNet}\label{overall-architecture-of-msvm-unet}
\begin{figure*}[ht]
	\centering
	\includegraphics[width=0.95\textwidth]{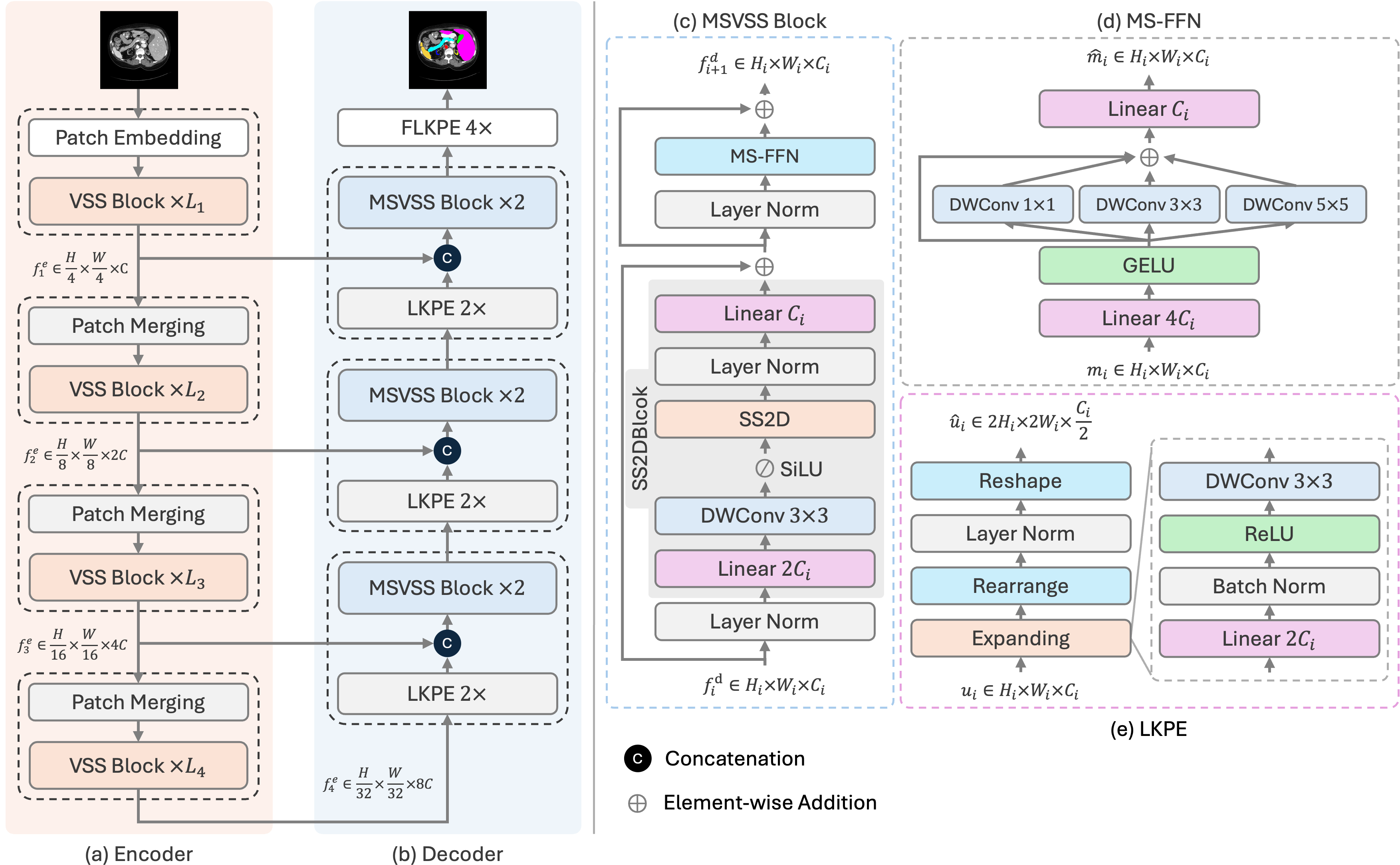}
	\caption{
		The overall architecture of our proposed MSVM-UNet. (a) The VMamba V2 encoder backbone network, (b) the decoder network composed of LKPE layers, MSVSS blocks, and FLKPE layers, (c) the Multi-Scale Visual State Space (MSVSS) block, (d) the Multi-Scale Feed-Forward Network (MS-FFN), and (e) the Large Kernel Patch Expanding (LKPE) layer. $f_1^e$, $f_2^e$, $f_3^e$, and $f_4^e$ are the output features of the four stages of hierarchical encoder backbones. $f_i^d$ and $f_{i + 1}^d$ represents the input and output features of the $i^{th}$ stage of the decoder, respectively.
	}
	\label{fig-1}
	\vspace{-\baselineskip}
\end{figure*}

The overall architecture of the proposed MSVM-UNet is illustrated in Fig.~\ref{fig-1}, which adopts a U-shaped hierarchical encoder-decoder structure with skip connections. The encoder is VMamba V2 \cite{liu2024vmamba} pre-trained on the ImageNet-1k dataset, which contains 4 stages. Except for the first stage consisting of a patch embedding layer and a VSS block, the other three stages are composed of a patch merging layer and a VSS block. Specifically, the patch embedding layer divides the input $X \in \mathbb{R}^{H \times W \times 3}$ into non-overlapping patches of size $4 \times 4$ and maps the channel dimension to dimension $C$. The VSS block is responsible for learning hierarchical feature representations of the input image. The patch merging layer is used for down-sampling the feature maps. For the input $X$, we first use the four stages of the encoder to sequentially extract four sets of hierarchical feature representations, denoted as $f_1^e$, $f_2^e$, $f_3^e$, and $f_4^e$, which are then fed into the decoder. Specifically, the feature $f_4^e$ is passed through the expanding path to the last stage of the decoder, while the features $f_1^e$, $f_2^e$, and $f_3^e$ are sent to the corresponding stages of the decoder via skip connections. The decoder consists of three stages, each comprising a Large Kernel Patch Expanding (LKPE) layer and a Multi-Scale Vision State Space (MSVSS) block. Unlike the patch merging layer, the LKPE layer is responsible for up-sampling the feature maps. The MSVSS block captures and aggregates fine-grained multi-scale information from the contracting path and high-level semantic information from the expanding path. Finally, the segmentation prediction is obtained through the final large kernel patch expanding (FLKPE) layer.

\subsection{Multi-Scale Vision State Space (MSVSS) Block}\label{multi-scale-vision-state-space-block}
To simultaneously capture multi-scale detail information in hierarchical features and effectively address the direction sensitivity issue in 2D visual data, we propose the Multi-Scale Visual State Space (MSVSS) block. Specifically, MSVSS addresses these challenges by introducing a Multi-Scale Feed-Forward Network (MS-FFN) within the VSS block. Firstly, the 2D-Selective-Scan Block (SS2DBlock) models the long-range dependencies of each feature in four directions, then the convolution operations in the MS-FFN aggregate information from the four remaining diagonal directions to enhance feature representation. Additionally, to effectively capture and aggregate multi-scale feature representations, the MSVSS employs a set of parallel convolution operations with different kernel sizes to achieve this goal. As shown in Fig.~\ref{fig-1}(c), the MSVSS block includes two layer normalization layers, an SS2DBlock, and an MS-FFN. The definition of the MSVSS block is given by Equations \eqref{eq-1} and \eqref{eq-2}:
\begin{equation}
	\hat{f}_i^d = SS2DBlock(LN(f_i^d)) + f_i^d
	\label{eq-1}
\end{equation}
\begin{equation}
	f_{i + 1}^d = MS\text{-}FFN(LN(\hat{f}_i^d)) + \hat{f}_i^d
	\label{eq-2}
\end{equation}
where $f_i^d$ and $f_{i + 1}^d$ denote the input and output feature maps of the $i^{th}$ stage, respectively. $\hat{f}_i^d$ represents the output of the SS2DBlock, and $LN(\cdot)$ denotes layer normalization.
\subsubsection{\textbf{2D-Selective-Scan Block (SS2DBlock)}}
\begin{figure*}[ht]
	\centering
	\includegraphics[width=0.95\textwidth]{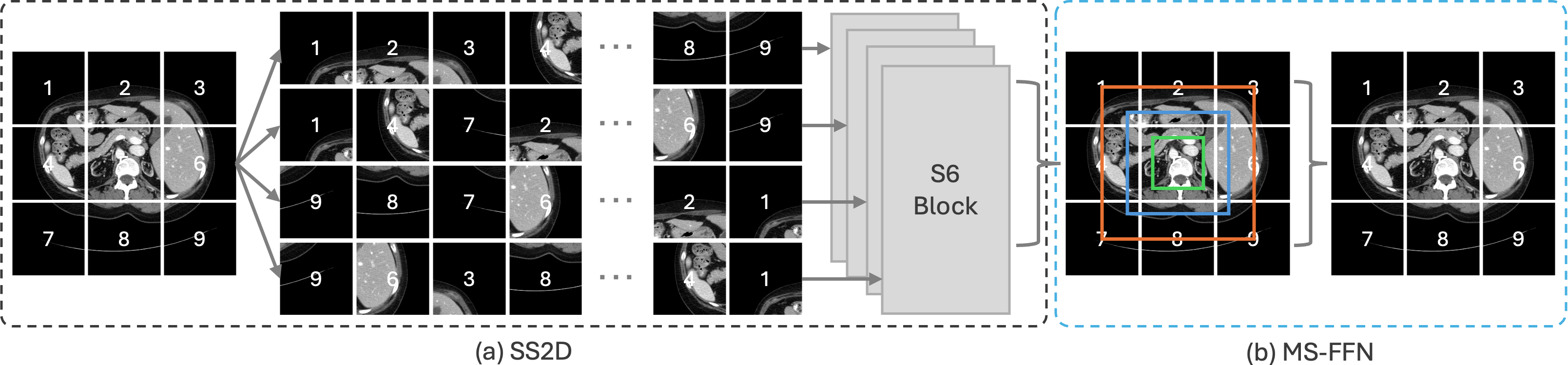}
	\caption{
		Illustration of the core operations in the MSVSS block. (a) The 2D-Selective-Scan (SS2D) operation. The input are first divided into patches and then flattened along four scanning paths, and then sent to S6 respectively. Finally, each of them is restored according to the scanning path and added together to obtain the output. (b) The Multi-scale Feed-Forward Neural Network (MS-FFN) layer. The input passes through multi-scale convolutions to further aggregate diagonal information and capture multi-scale information representation.
	}
	\label{fig-2}
	\vspace{-\baselineskip}
\end{figure*}

2D-Selective-Scan block performs selective scanning along four scanning paths on the input feature map to capture global contextual information and long-range dependencies. Specifically, the 2D input feature map first undergoes a linear layer, a depth-wise convolution operation, and an activation function. Then, further feature extraction is carried out through the 2D-Selective-Scan (SS2D) operation. Finally, the output is obtained after another layer normalization and a linear projection. As shown in Fig.~\ref{fig-2}(a), SS2D first flattens the 2D input along four different scanning paths to obtain four 1D sequences. These sequences are then fed into the S6 blocks \cite{gu2023mamba} for selective scanning to model long-range dependencies. Finally, the four 1D sequences are restored to their original 2D form and summed to produce the output. The definition of the SS2DBlock is given by Equation \eqref{eq-3}:
\begin{equation}
	\hat{z}_i = C_2(LN(SS2D(\sigma_1(DWConv(C_1(z_i))))))
	\label{eq-3}
\end{equation}
where $z_i$ and $\hat{z}_i$ represent the input and output feature maps of the SS2DBlock in the $i^{th}$ stage, respectively. $C_1(\cdot)$ represents a linear projection used to double the channel dimension. $DWConv(\cdot)$ denotes a depth-wise convolution with a kernel size of $3 \times 3$. $\sigma_1$ represents a $SiLU(x) = x \cdot sigmoid(x)$ activation function. $SS2D(\cdot)$ denotes 2D-Selective-Scan operation. $C_2(\cdot)$ represents another linear projection used to reduce the channel dimension by half.

\subsubsection{\textbf{Multi-Scale Feed-Forward Neural Network (MS-FFN)}}
As depicted in Fig.~\ref{fig-2}(b), we introduce convolution operations in the feed-forward network to aggregate information from these four diagonal directions. Additionally, to effectively capture detailed information and multi-scale feature representations in hierarchical features, we employ a set of convolution operations with different kernel sizes. To avoid introducing excessive parameters and computational overhead, we use depth-wise convolutions, which are parameter and computation efficient, to implement the MS-FFN. As shown in Fig.~\ref{fig-1}(d), the MS-FFN consists of two linear layers and a set of parallel depth-wise convolution layers. The definitions of MS-FFN are given by Equations \eqref{eq-4} and \eqref{eq-5}:
\begin{equation}
	m_i^{'} = \sigma_2(C_3(m_i))
	\label{eq-4}
\end{equation}
\begin{equation}
	\hat{m}_i = C_4(\sum_{ks \in KS} DWConv_{ks}(m_i^{'}) + m_i^{'})
	\label{eq-5}
\end{equation}
where $m_i$ and $\hat{m}_i$ represent the input and output tensors of the MS-FFN in the $i^{th}$ stage, respectively, and $m_i^{'}$ denote the output of the first linear transformation. $C_3(\cdot)$ represents a linear projection used to expand the channel dimension by four times. $DWConv_{ks}(\cdot)$ denotes a depth-wise convolution with a kernel size of $ks$. $KS$ defines a group of parallel convolution kernels with values of $\{1 \times 1, 3 \times 3, 5 \times 5 \}$.  $\sigma_2(\cdot)$ represents a $GELU(x) = x \cdot \Phi(x)$ activation function. 
$C_4(\cdot)$ represents another linear projection that reduces the channel dimension back to the input dimension.

\subsection{Large Kernel Patch Expanding (LKPE) Layer}\label{large-kernel-patch-expanding-layer}
We use the LKPE layer to upsample the feature maps of the current stage to match the dimensions of the feature maps from the skip connection. As shown in Fig.~\ref{fig-3}, it compares our proposed LKPE layer with the patch expanding layer proposed by Swin-UNet. Unlike the latter method, which relies solely on linear projections (equivalent to convolutions with a kernel size of $1 \times 1$) for expanding the input feature channel dimension, we consider introducing large convolution kernels. Inspired by other upsampling methods, such as transposed convolutions and UpSample \cite{rahman2023medical}, the patch expanding layer only considers the channel information of features and neglects the spatial relationships between adjacent features, making this approach suboptimal in terms of information utilization. To solve this problem, we propose the large kernel patch expanding layer. Specifically, LKPE first applies a $1 \times 1$ convolution to double the channel dimension, followed by batch normalization and a ReLU activation function. Next, it uses an efficient depth-wise convolution to aggregate spatial information, and finally performs upsampling by expanding the feature representation that integrates both spatial and channel information. The definition of LKPE is given by Equation \eqref{eq-6}:
\begin{figure}[th]
	\centering
	\includegraphics[width=.48\textwidth]{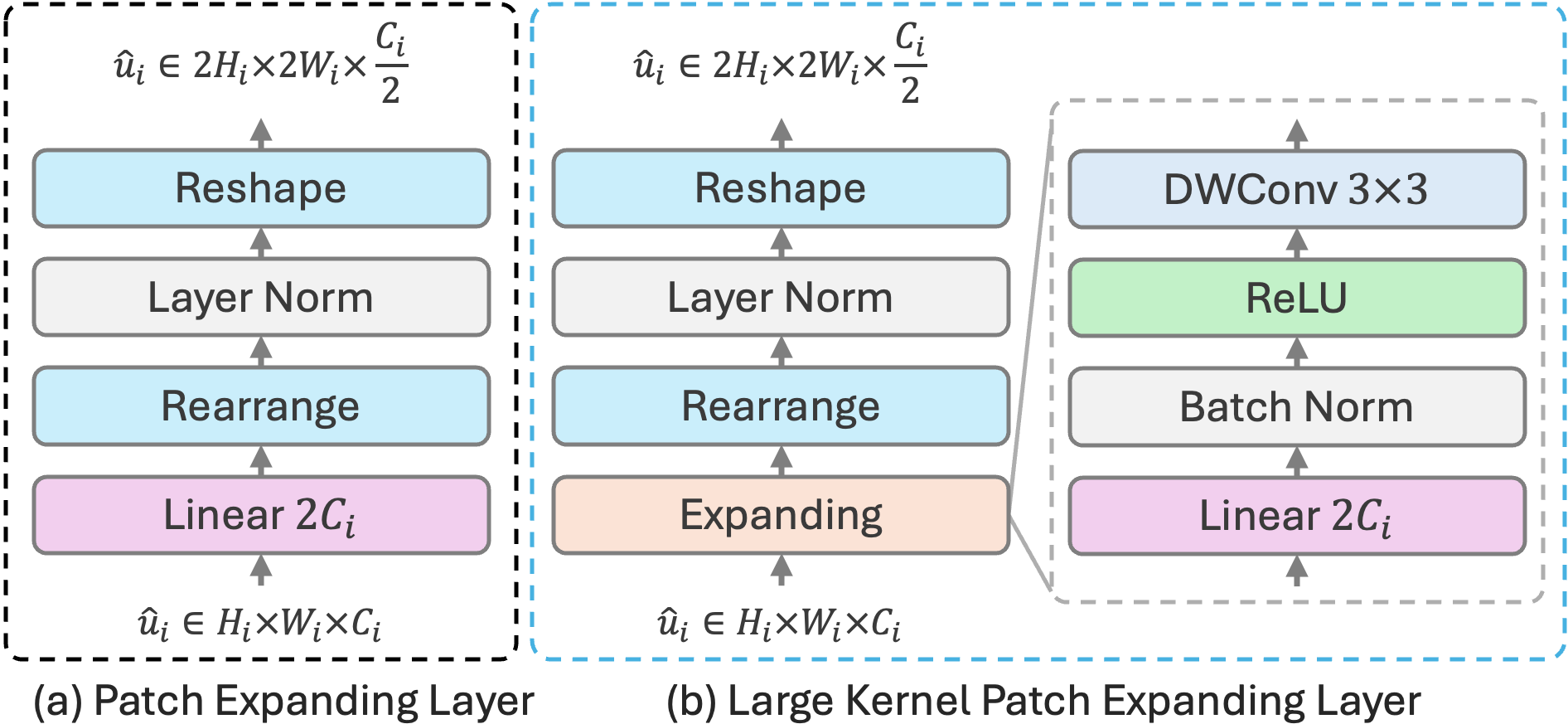}
	\caption{
		Comparison of different patch expanding layers. (a) The patch expanding layer proposed by Swin-UNet, (b) the large kernel patch expanding (LKPE) layer proposed by us.
	}
	\label{fig-3}
	\vspace{-\baselineskip}
\end{figure}
\begin{equation}
	\hat{u}_i = RS(LN(RA(DWConv(\sigma_3(BN(C_5(u_i)))))))
	\label{eq-6}
\end{equation}
where $u_i$ and $\hat{u}_i$ represent the feature maps before and after upsampling at the $i^{th}$ stage, respectively. $C_5(\cdot)$ represents a linear transformation responsible for doubling the channel dimension. $BN(\cdot)$ represents batch normalization. $\sigma_3(\cdot)$ denotes a $ReLU(x) = max(0, x)$ activation function. $RA(\cdot)$ denotes the rearrange operation, and $RS(\cdot)$ denotes the reshape operation.

\begin{table*}[htbp]
	\centering
	\caption{
		Performance comparison with state-of-the-art methods on the Synapse multi-organ dataset. Bold black data indicates the best results, and underlined black data denotes the second-best results.
	}
	\begin{tabular*}{\textwidth}{@{\extracolsep{\fill}}c|cc|cccccccc@{\extracolsep{\fill}}}
		\toprule
		Methods & DSC (\%) $\uparrow$ & HD95 (mm) $\downarrow$ & Aorta & Gallbladder & Kidney (L) & Kidney (R) & Liver & Pancreas & Spleen & Stomach \\
		\midrule
		UNet \cite{ronneberger2015u} & 74.82 & 54.59 & 85.66 & 53.24  & 81.13 & 71.60 & 92.69 & 56.81 & 87.46 & 69.93 \\
		Att-UNet \cite{rahman2023medical} & 71.70 & 34.47 & 82.61 & 61.94 & 76.07 & 70.42 & 87.54 & 46.70 & 80.67 & 67.66 \\
		TransUNet \cite{chen2021transunet} & 76.76 & 44.31 & 86.71 & 58.97 & 83.33 & 77.95 & 94.13 & 53.60 & 84.00 & 75.38 \\
		MISSFormer \cite{huang2021missformer} & 81.96 & 18.20 & 86.99 & 68.65 & 85.21 & 82.00 & 94.41 & 65.67 & 91.92 & 80.81 \\
		Swin-UNet \cite{cao2022swin} & 79.13 & 21.55 & 85.47 & 66.53 & 83.28 & 79.61 & 94.29 & 56.58 & 90.66 & 76.60 \\
		PVT-CASCADE \cite{rahman2023medical} & 81.06 & 20.23 & 83.01 & 70.59 & 82.23 & 80.37 & 94.08 & 64.43 & 90.10 & 83.69 \\
		TransCASCADE \cite{rahman2023medical} & 82.68 & 17.34 & 86.63 & 68.48 & 87.66 & 84.56 & 94.43 & 65.33 & 90.79 & 83.52 \\
		2D D-LKA Net \cite{azad2024beyond} & 84.27 & 20.04 & \underline{88.34} & 73.79 & \textbf{88.38} & \textbf{84.92} & 94.88 & 67.71 & 91.22 & \underline{84.94} \\
		MERIT-GCASCADE \cite{rahman2024g} & \underline{84.54} & \textbf{10.38} & 88.05 & \underline{74.81} & 88.01 & \underline{84.83} & \underline{95.38} & \underline{69.73} & 91.92 & 83.63 \\
		PVT-EMCAD-B2 \cite{rahman2024emcad} & 83.63 & 15.68 & 88.14 & 68.87 & \underline{88.08} & 84.10 & 95.26 & 68.51 & \underline{92.17} & 83.92 \\
		VM-UNet \cite{ruan2024vm} & 82.38 & 16.22 & 87.00 & 69.37 & 85.52 & 82.25 & 94.10 & 65.77 & 91.54 & 83.51 \\
		Swin-UMamba \cite{liu2024swin} & 82.26 & 19.51 & 86.32 & 70.77 & 83.66 & 81.60 & 95.23 & 69.36 & 89.95 & 81.14 \\
		\midrule
		MSVM-UNet (ours) & \textbf{85.00} & \underline{14.75} & \textbf{88.73} & \textbf{74.90} & 85.62 & 84.47 & \textbf{95.74} & \textbf{71.53} & \textbf{92.52} & \textbf{86.51} \\
		\bottomrule
	\end{tabular*}
	\label{tab-1}
	\vspace{-\baselineskip}
\end{table*}

\subsection{Final Large Kernel Patch Expanding (FLKPE) Layer}\label{final-large-kernel-patch-expanding-layer}
We use the FLKPE layer to generate the segmentation prediction. Taking the feature map from the final stage of the decoder as input, we first apply a linear projection to aggregate the channel dimension information and expand the channel dimension by 16 times. Next, we use a depth-wise convolution to aggregate spatial dimension information. Subsequently, we transform the spatial resolution of the resulting features to match the size of the input image while keeping the channel dimension unchanged. Finally, the transformed feature map is passed through a $1 \times 1$ convolution to map it to the segmentation prediction. The definition of FLKPE is given by Equation \eqref{eq-7}:
\begin{equation}
	\begin{gathered}
		\hat{f}_3^d = RS(LN(RA(DWConv(\sigma_3(BN(C_6(f_3^d))))))) \\
		y = Conv_{1 \times 1}(\hat{f}_3^d)
	\end{gathered}
	\label{eq-7}
\end{equation}
where $f_3^d$ represents the feature map output by the last stage of the decoder. $\hat{f}_3^d$ represents the feature map obtained by upsampling $f_3^d$ to the input image size. $y$ denotes the final segmentation prediction. $C_6(\cdot)$ represents a linear transformation responsible for expanding the input channel dimension by 16 times.

\section{EXPERIMENTS AND RESULTS}
\subsection{Datasets}\label{datasets}
We evaluate the performance of the proposed MSVM-UNet on the Synapse abdominal multi-organ segmentation dataset (Synapse) and the Automated Cardiac Diagnosis Challenge dataset (ACDC).
\subsubsection{\textbf{Synapse dataset} \cite{landman2015miccai}}
The dataset consists of 30 abdominal CT scans with 3779 axial contrast-enhanced abdominal CT images. Each CT volume contains 85 to 198 slices of $512 \times 512$ pixels, with a voxel spatial resolution of ($[0.54 \sim 0.54] \times [0.98 \sim 0.98] \times [2.5 \sim 5.0]$) $mm^3$. Similar to TransUNet \cite{chen2021transunet}, we randomly divide the dataset into 18 cases for training and 12 cases for testing. We segment only 8 types of abdominal organs: aorta, gallbladder, left kidney, right kidney, liver, pancreas, spleen, and stomach.
\subsubsection{\textbf{ACDC dataset} \cite{bernard2018deep}}
The dataset contains 100 cardiac MRI scan images, each of which consists of three sub-organs: right ventricle (RV), myocardium (Myo), and left ventricle (LV). Following the TransUNet \cite{chen2021transunet}, we split the data into 70 cases for training, 10 for validation, and 20 for testing. \hspace{5em}

\subsection{Implementation Details and Evaluation Metrics}\label{implementation-details-and-evaluation-metrics}
\subsubsection{\textbf{Implementation Details}}
In our experiments, all models were implemented based on the Pytorch 2.0.0 framework, and all training was conducted on an NVIDIA GeForce RTX 3090 GPU. We initialized the backbone network using pretrained weights from the ImageNet-1k dataset. To reduce overfitting and enhance the model's generalization capability, we employed extensive data augmentation techniques, including resizing the input images to $224 \times 224$, horizontal and vertical flips, random rotations, Gaussian noise, Gaussian blur, and contrast enhancement. We set the batch size to 32 and used the AdamW optimizer to train the network for a maximum of 300 epochs. The initial learning rate was set to 5e-4 and was decayed during training using a cosine annealing schedule. Due to the varying difficulty levels of different datasets and to reduce overfitting, we set different weight decay for different datasets: 1e-3 for the Synapse dataset and 1e-4 for the ACDC dataset. Additionally, the reported FLOPs and parameter counts of the models were calculated using calflops with an input size of $224 \times 224 \times 3$. We used a combination of Dice and cross-entropy loss functions to train the network, defined as follows:
\begin{equation}
	L_{total} = \alpha L_{DICE} + (1 - \alpha) L_{CE}
	\label{eq-8}
\end{equation}
where $\alpha = 0.6$ and $1 - \alpha = 0.4$ are the weights for the Dice loss $L_{DICE}$ and the cross-entropy loss $L_{CE}$, respectively. \hspace{2em}
\subsubsection{\textbf{Evaluation Metrics}} 
Following the commonly adopted metrics to measure model performance,  we used the Dice SCore (DSC) and the 95\% Hausdorff Distance (HD95) to assess our model's performance on the Synapse and ACDC datasets. The DSC and HD are calculated according to Equations \eqref{eq-9} and \eqref{eq-10}:
\begin{equation}
	DSC(X, Y) = \frac{2 * \vert X \cap Y \vert}{\vert X \vert + \vert Y \vert}
	\label{eq-9}
\end{equation}
\begin{equation}
	\begin{gathered}
		HD(X, Y) = \mathop{max}\{h(X, Y), h(Y, X)\} \\
		h(X, Y) = \mathop{max}_{x \in X} \mathop{min}_{y \in Y} d(x, y) \\
		h(Y, X) = \mathop{max}_{y \in Y} \mathop{min}_{x \in X} d(x, y)
	\end{gathered}
	\label{eq-10}
\end{equation}
where the $X$ and $Y$ denote the ground truth and segmented maps, respectively, and $d(x, y)$ represents the distance between points $x$ and $y$. HD95 is the $95^{th}$ percentile of the distances of the boundaries of $X$ and $Y$.

\begin{figure*}[ht]
	\centering
	\includegraphics[width=0.95\textwidth]{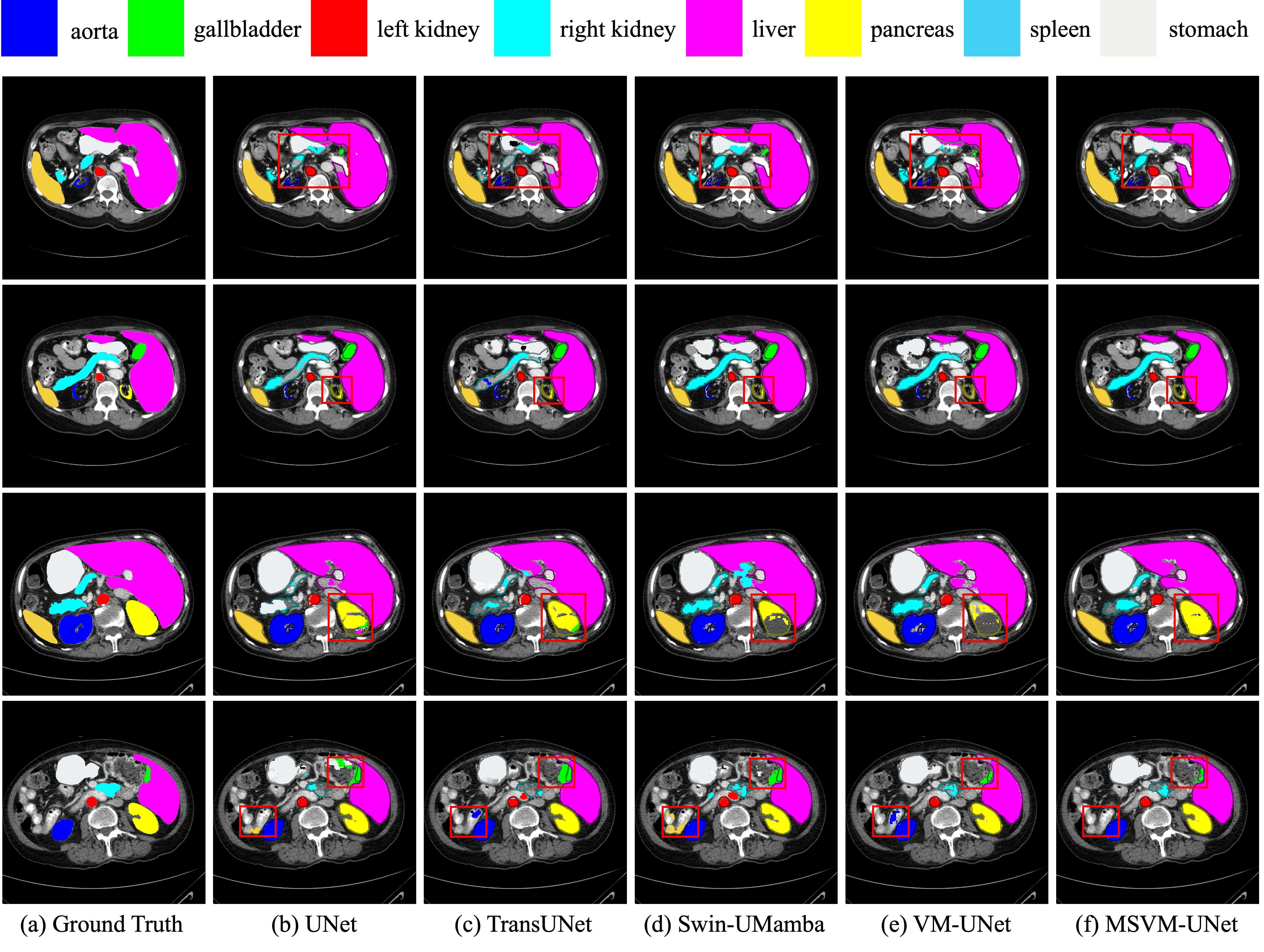}
	\caption{
		Visual comparison of different methods on the Synapse multi-organ dataset. The first column represents the ground truth, and the following columns represent the segmentation predictions of UNet, TransUNet, Swin-UMamba, VM-UNet, and MSVM-UNet methods, respectively. The superiority of our proposed method can be clearly seen in the organ regions highlighted by red rectangles. Various colored contour lines indicate the ground truth for the corresponding organs.
	}
	\label{fig-4}
	\vspace{-\baselineskip}
\end{figure*}

\subsection{Comparisons with State-of-the-Arts}\label{comparisons-with-state-of-the-arts}
To verify the effectiveness of our proposed method, we compared its performance with state-of-the-art methods based on CNN, transformer, and mamba.
\subsubsection{\textbf{Results on Synapse Multi-Organ Segmentation}}
As shown in Table~\ref{tab-1}, compared to various types of methods, our proposed MSVM-UNet achieved the best average DSC of 85.00\% and HD95 of 14.75mm. Specifically, compared to CNN-based methods (such as 2D D-LKA Net), our method improved the DSC and HD95 by 0.73\% and 5.29mm, respectively; by 1.37\% and 0.93mm compared to transformer-based methods (such as PVT-EMCAD-B2); and by 2.62\% and 1.47mm compared to mamba-based methods (such as VM-UNet). Additionally, for the small organs, there were 0.09\% and 1.80\% DSC improvements in the gallbladder and pancreas, respectively, compared to the second-best methods, and a 1.57\% improvement in the stomach for the large organ. This is because MSVM-UNet can effectively capture both long-range dependencies between pixels and local contextual relationships simultaneously. Due to the introduction of multi-scale convolution operations, MSVM-UNet not only handles organs of varying shapes and sizes effectively but also better locates organ boundaries.
\subsubsection{\textbf{Results on ACDC for Automated Cardiac Segmentation}}
As shown in Table~\ref{tab-2}, we present the performance of our method compared to the aforementioned types of methods on MRI medical images, where our proposed MSVM-UNet achieved the best average DSC of 92.58\%. Additionally, for the three categories in the ACDC dataset (RV, Myo, and LV), our method achieved the best DSC results of 91.00\%, 90.35\%, and 96.39\%, respectively. This demonstrates the generalizability of our method, as it performs well on different modalities of medical image data (MRI and CT).

\begin{table}[t]
	\caption{
		Performance comparison with state-of-the-art methods on the ACDC dataset. Bold black data indicates the best results, and underlined black data denotes the second-best results.
	}
	\begin{tabular*}{\columnwidth}{c|c|@{\extracolsep{\fill}}ccc}
		\toprule
		Methods & DSC (\%) $\uparrow$ & RV & Myo & LV \\
		\midrule
		R50 UNet \cite{chen2021transunet} & 87.55 & 87.10 & 80.63 & 94.92 \\
		R50 Att-UNet \cite{chen2021transunet} & 86.75 & 87.58 & 79.20 & 93.47 \\
		TransUNet \cite{chen2021transunet} & 89.71 & 88.86 & 84.53 & 95.73 \\
		Swin-UNet \cite{cao2022swin} & 90.00 & 88.55 & 85.62 & 95.83 \\
		MISSFormer \cite{huang2021missformer} & 90.86 & 89.55 & 88.04 & 94.99 \\
		PVT-CASCADE \cite{rahman2023medical} & 91.46 & 88.90 & 89.97 & 95.50 \\
		TransCASCADE \cite{rahman2023medical} & 91.63 & 89.14 & \underline{90.25} & 95.50 \\
		MERIT-GCASCADE \cite{rahman2024g} & 92.23 & 90.64 & 89.96 & \underline{96.08} \\
		PVT-EMCAD-B2 \cite{rahman2024emcad} & 92.12 & 90.65 & 89.68 & 96.02 \\
		VM-UNet \cite{ruan2024vm} & \underline{92.24} & 90.74 & 89.93 & 96.03 \\
		Swin-UMamba \cite{liu2024swin} & 92.14 & \underline{90.90} & 89.80 & 95.72 \\
		\midrule
		MSVM-UNet (ours) & \textbf{92.58} & \textbf{91.00} & \textbf{90.35} & \textbf{96.39} \\
		\bottomrule
	\end{tabular*}
	\label{tab-2}
	\vspace{-\baselineskip}
\end{table}

\subsection{Qualitative Analysis}\label{aualitative-analysis}
As shown in Fig.~\ref{fig-4}, we present a 2D visual comparison of our method with other methods on the Synapse multi-organ dataset. It can be observed that our method produces better segmentation results across different organs and, to some extent, avoids issues of over-segmentation (as seen in the segmentation of the gallbladder in the last row) and under-segmentation (as seen in the segmentation of the pancreas in the third row). This is because our MSVM-UNet can better capture features with varying geometric shapes. Compared to mamba-based methods, our approach shows better segmentation performance for organ boundaries. Additionally, when comparing the liver segmentation in the first row, we found that methods incorporating convolution operations achieve better results in delineating the liver boundary compared to methods without convolution operations. This is due to the inclusion of appropriate convolution operations, which help the model capture local detail features and positional information, leading to more discriminative feature representations and better segmentation results.

\begin{table}[t]
	\tabcolsep = 4pt 
	\caption{
		Ablation study on the effect of different components of MSVM-UNet with VMamba V2 Tiny encoder on the Synapse multi-organ dataset.
	}
	\begin{tabular*}{\columnwidth}{@{\extracolsep{\fill}}cc|cccc@{\extracolsep{\fill}}}
		\toprule
		\multicolumn{2}{c|}{Components} & \multirow{2}{*}{DSC (\%)} & \multirow{2}{*}{HD95 (mm)} & \multirow{2}{*}{\#FLOPs (G)} & \multirow{2}{*}{\#Params (M)} \\
		\cline{1-2} \raisebox{-0.5\height}[0pt]{MSVSS} & \raisebox{-0.5\height}[0pt]{LKPE} & & & & \\
		\midrule
		No & No & 82.80 & 27.29 & \textbf{15.11} & \textbf{35.68} \\
		Yes & No & 84.67 & 15.83 & 15.40 & 35.87 \\
		No & Yes & 84.30 & 14.21 & 15.23 & 35.74 \\
		Yes & Yes & \textbf{85.00} & \textbf{14.75} & 15.53 & 35.93 \\
		\bottomrule
	\end{tabular*}
	\label{tab-3}
	\vspace{-\baselineskip}	
\end{table}

\subsection{Ablation Studies}\label{ablation-studies}
We conducted a comprehensive ablation study on the Synapse dataset to validate and investigate the effectiveness of our proposed method and to demonstrate the rationale behind our architectural design choices. Unless otherwise specified, all ablation experiments use the tiny version of VMamba V2 pre-trained on the ImageNet-1k as the encoder.
\subsubsection{\textbf{Effect of Different Components of MSVM-UNet}}
We conducted a series of experiments on the Synapse multi-organ dataset to understand the impact of different components in the MSVM-UNet decoder. We assessed the effects of replacing specific modules in the decoder with our proposed modules. As shown in Table~\ref{tab-3}, significant performance improvements were observed after substituting the original modules with our proposed ones, with only a minimal increase in computational overhead and the number of parameters. Specifically, compared to the decoder with VSS blocks and patch expanding layers, using our proposed modules improved the DSC and HD95 by 2.2\% and 12.54mm, respectively, with only an additional 0.42G FLOPs and 0.25M parameters. This demonstrates both the effectiveness and efficiency of our proposed MSVM-UNet.

\begin{table}[t]
	\tabcolsep = 4pt
	\caption{
		Ablation study on the effect of different upsampling methods on the Synapse multi-organ dataset.
	}
	\begin{tabular*}{\columnwidth}{@{\extracolsep{\fill}}c|cccc@{\extracolsep{\fill}}}
		\toprule
		Upsampling & DSC (\%) & HD95 (mm) & \#FLOPs (G) & \#Params (M) \\
		\midrule
		Transposed Conv & 83.14 & \textbf{13.84} & \textbf{4.45} & \textbf{6.06} \\
		UpSample \cite{rahman2023medical} & 83.94 & 16.65 & 7.22 & 8.00 \\
		Patch Expand \cite{cao2022swin} & 82.80 & 27.29 & 5.44 & 6.20 \\
		LKPE (ours) & \textbf{84.30} & 14.21 & 5.56 & 6.26 \\
		\bottomrule
	\end{tabular*}
	\label{tab-4}
	\vspace{-\baselineskip}	
\end{table}

\subsubsection{\textbf{Effect of Different Upsampling Methods}}
To explore the effectiveness of our proposed LKPE, we conducted experiments on the Synapse multi-organ dataset by using the original decoder with transposed convolution, the UpSample block  \cite{rahman2023medical}, the patch expanding layer, and the LKPE layer as upsampling layers, respectively, to evaluate their individual performance. As shown in Table~\ref{tab-4}, we reported the performances and overheads corresponding to different upsampling methods. To provide a clearer comparison of the computational overhead and the number of parameters of different upsampling operations, we only report the FLOPs and parameters of the decoder. Compared to the original patch expanding layer, LKPE improved the DSC and HD95 by 1.5\% and 13.08mm, respectively, while only introducing an additional 0.12G FLOPs and 0.06M parameters. Moreover, it can be observed that methods aggregating channel and spatial information generally achieve better performance, further demonstrating the effectiveness of our proposed LKPE.

\begin{table}[t]
	\caption{
		Ablation study on the effect of multi-scale kernels in depth-wise convolution of the MSVSS Block on the Synapse multi-organ dataset.
	}
	\begin{tabular*}{\columnwidth}{@{\extracolsep{\fill}}c|cccc@{\extracolsep{\fill}}}
		\toprule
		Conv. kernels & DSC (\%) & HD95 (mm) & \#FLOPs (G) & \#Params (M) \\
		\midrule
		$[1, 3, 5]$ & \textbf{84.67} & \textbf{15.83} & \textbf{5.73} & \textbf{6.39} \\
		$[3, 5, 7]$ & 84.45 & 18.65 & 6.14 & 6.65 \\
		$[1, 3, 5, 7]$ & 84.40 & 18.33 & 6.14 & 6.66 \\
		\bottomrule
	\end{tabular*}
	\label{tab-5}
	\vspace{-\baselineskip}
\end{table}

\subsubsection{\textbf{Effect of Multi-Scale Kernels in MSVSS Block}}
We also conducted additional experiments on the Synapse multi-organ dataset to explore the impact of different multi-scale convolution kernels in the depth-wise convolution of the MSVSS block. As shown in Table~\ref{tab-5}, the performances for various multi-scale convolution kernels are reported. Similarly, the FLOPs and the number of parameters are provided only for the decoder. To avoid excessive computational overhead, we designed three sets of kernels: $1 \times 1$ and $7 \times 7$ in the first set, $3 \times 3$ and $5 \times 5$ in the second, and a union of both in the third. As the number and size of convolution kernels increase, we found that the performance decreases. Based on these observations, we selected $[1, 3, 5]$ as the default multi-scale convolution kernels in the MSVSS block.

\begin{table}[t]
	\caption{
		Ablation study on the effect of the encoder model scales on the Synapse multi-organ dataset.
	}
	\begin{tabular*}{\columnwidth}{@{\extracolsep{\fill}}c|cccc@{\extracolsep{\fill}}}
		\toprule
		Encoder scales & DSC (\%) & HD95 (mm) & \#FLOPs (G) & \#Params (M) \\
		\midrule
		tiny & \textbf{85.00} & 14.75 & \textbf{15.53} & \textbf{35.93} \\
		small & 84.75 & \textbf{14.66} & 22.86 & 54.69 \\
		\bottomrule
	\end{tabular*}
	\label{tab-6}
	\vspace{-\baselineskip}	
\end{table}

\subsubsection{\textbf{Effect of Encoder Model Scales}}
To investigate the impact of different encoder depths on model performance, we conducted two sets of experiments on the Synapse multi-organ dataset to study the effects of varying encoder scales. As shown in Table~\ref{tab-6}, there is a slight decrease in performance as the encoder becomes larger and deeper.  Since both sets of experiments used the same setup, we hypothesize that this minor performance drop may be due to slight overfitting caused by the increased complexity of the model. Based on these observations and considering the computational overhead and the number of parameters, we chose the tiny version of the encoder as the default scale.

\begin{figure}[t]
	\centering
	\includegraphics[width=.48\textwidth]{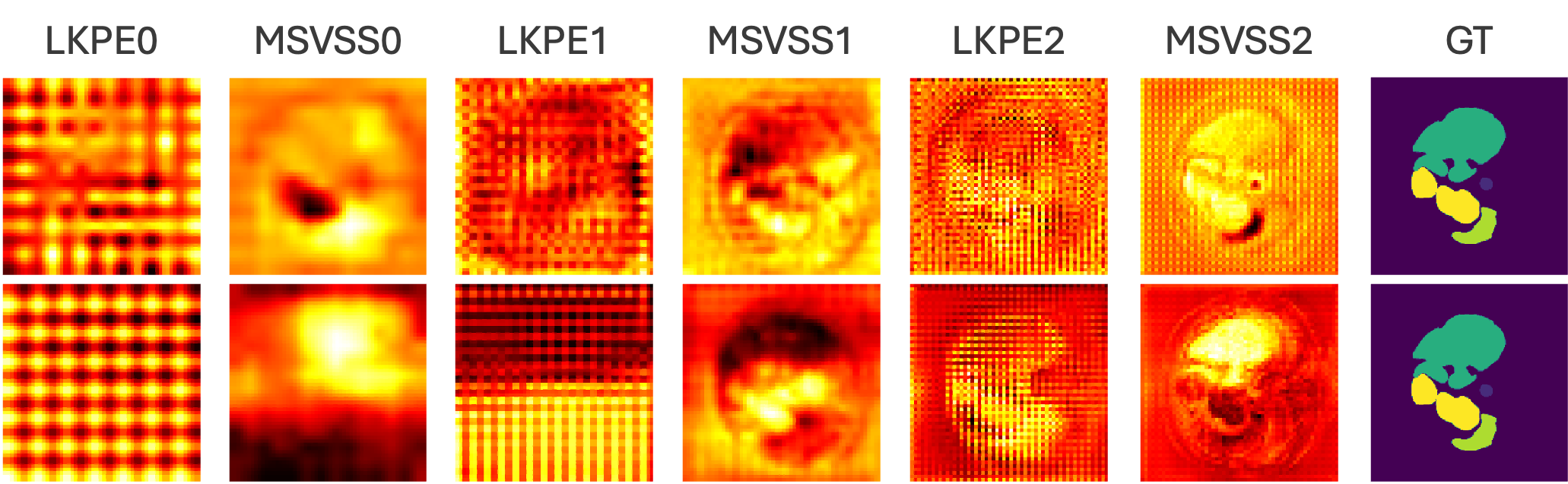}
	\caption{
		Visual comparison of decoder features with and without our proposed blocks. The first and second rows present the features of our decoder (with our proposed blocks) and the original decoder (without our proposed blocks), respectively. Layer numbers are numbered from the bottom to the top of the decoder.
	}
	\label{fig-5}
	\vspace{-\baselineskip}
\end{figure}

\subsubsection{\textbf{Effect of Feature Enhancement}}
The features of the corresponding layers of the original decoder and our MSVM-UNet decoder are visualized in Fig.~\ref{fig-5}. We compute the average of all channels in the feature map and then produce the heatmap using Matplotlib. It is evident from Fig.~\ref{fig-5} that our method helps in handling organs of varying sizes and shapes and in obtaining more discriminative feature representations.

\section{CONCLUSION}
In this paper, we propose a novel multi-scale visual Mamba UNet to address medical image segmentation challenges. Thanks to the design of multi-scale depth-wise convolution, MSVM-UNet not only captures information at various scales and models long-range dependencies in all directions but also maintains computational efficiency and acceptable parameter counts. Additionally, by effectively integrating channel and spatial information for upsampling, MSVM-UNet achieves more discriminative feature representations, leading to more accurate medical image segmentation results. Our experimental results demonstrate the outstanding performance of MSVM-UNet on two medical image datasets, with improvements in DSC and HD95 on the Synapse multi-organ dataset surpassing VM-UNet by 2.62\% and 1.47mm, respectively. Qualitative analysis also shows that MSVM-UNet can accurately localize organs and handle organs of varying sizes and shapes.

\section*{Acknowledgment}
This work was supported by the National Natural Science Foundation of China (62062067).

\small
\bibliography{references.bib}
\bibliographystyle{IEEEtran}

\end{document}